\documentclass[sigconf]{acmart}
\AtBeginDocument{%
  \providecommand\BibTeX{{%
    \normalfont B\kern-0.5em{\scshape i\kern-0.25em b}\kern-0.8em\TeX}}}

\setcopyright{acmcopyright}
\copyrightyear{2021}
\acmYear{2021}
\acmDOI{10.1145/1122445.1122456}

\acmConference[KDD Health Day / DSHealth 2021]{KDD Health Day / DSHealth 2021: Joint KDD 2021 Health Day and 2021 KDD Workshop on Applied Data Science for Healthcare}{August 14--18, 2021}{Virtual}
\acmBooktitle{KDD Health Day / DSHealth 2021: Joint KDD 2021 Health Day and 2021 KDD Workshop on Applied Data Science for Healthcare, August 14--18, 2021, Virtual}
\acmPrice{15.00}
\acmISBN{978-1-4503-XXXX-X/18/06}

\begin{document}

\title{Quantifying machine learning-induced overdiagnosis in sepsis}

\author{Anna Fedyukova}
\authornotemark[2]
\email{afedyukova@student.unimelb.edu.au}
\orcid{0000-0002-1861-812X}
\affiliation{%
  \institution{The University of Melbourne}
  \streetaddress{Parkville}
  \city{Melbourne}
  \state{VIC}
  \country{Australia}
  \postcode{3010}
 }

\author{Douglas Pires}
\authornotemark[1]
\email{douglas.pires@unimelb.edu.au}
\orcid{0000-0002-3004-2119}
\affiliation{%
  \institution{The University of Melbourne}
  \streetaddress{Parkville}
  \city{Melbourne}
  \state{VIC}
  \country{Australia}
  \postcode{3010}
}

\author{Daniel Capurro}
\authornote{Authors contributed equally to this research.}
\email{dcapurro@unimelb.edu.au}
\orcid{0000-0002-9256-1256}
\affiliation{%
  \institution{The University of Melbourne}
  \streetaddress{Parkville}
  \city{Melbourne}
  \state{VIC}
  \country{Australia}
  \postcode{3010}
}

\renewcommand{\shortauthors}{Fedyukova et al.}

\begin{abstract}
The proliferation of early diagnostic technologies, including self-monitoring systems and wearables, coupled with the application of these technologies on large segments of healthy populations may significantly aggravate the problem of \textit{overdiagnosis}. This can lead to unwanted consequences such as overloading health care systems and overtreatment, with potential harms to healthy individuals. The advent of machine-learning tools to assist diagnosis---while promising rapid and more personalised patient management and screening---might contribute to this issue.

The identification of overdiagnosis is usually \textit{post hoc} and demonstrated after long periods (from years to decades) and costly randomised control trials. In this paper, we present an innovative approach that allows us to preemptively detect potential cases of overdiagnosis during predictive model development. This approach is based on the combination of labels obtained from a prediction model and clustered medical trajectories, using sepsis in adults as a test case. This is one of the first attempts to quantify machine-learning induced overdiagnosis and we believe will serves as a platform for further development, leading to guidelines for safe deployment of computational diagnostic tools.
\end{abstract}




\maketitle

\section{Introduction}
Overdiagnosis consists of identifying abnormalities that meet disease definitions  but will never cause symptoms or death during a patient`s ordinarily expected lifetime~\cite{ref1}. It causes significant harm and increases costs for healthcare systems~\cite{ref2}. The drawbacks of overdiagnosis are unnecessary labelling, adverse effects of additional health examinations and treatments, long term-medication intake, anxiety, and waste of resources which are needed to treat or prevent genuine illnesses~\cite{ref1, ref2, ref3}. According to a previous study, from 24.4\% to 48.3\% of women aged 35 to 84 years were overdiagnosed with breast cancer (including ductal carcinoma \textit{in situ}) in 2010~\cite{ref9}. Another work shows that 30.2\% of individuals, participating in a study and diagnosed with asthma, did not present the condition and, among them, 65.5\% did not need any medical treatment~\cite{ref10}.

There are two main scenarios where overdiagnosis can take place: (i) overdetection of mild “abnormalities” that likely will never compromise a patient's life~\cite{ref1} and (ii) overdefinition – inclusion of patients with minor or uncertain symptoms by broadening definitions and lowering thresholds for risk factors~\cite{ref2, ref3}.  The increased availability of digital patient data and machine learning models provide a perfect substrate for overdiagnosis due to several reasons. Firstly, the definition of a disease is translated by a developer to an algorithm. Secondly, access to fast and cheap diagnostic tools might induce its broader application by clinicians, increasing chances for overdiagnosis. Thirdly, combining diagnostic algorithms and self-trackers and wearables leads to application of these technologies to broader and healthier segments of the population, aggravating the risk of overdiagnosis. 

Overdiagnosis is usually observed \textit{post hoc}, during randomised controlled trials, with the majority of efforts taking from years or decades, when the harm to patients has already been done.

The purpose of this study is to propose novel methods to preemptively estimate the proportion of overdiagnosed patients generated by a new digital diagnostic algorithm, as means to mitigate these during development stages.

\begin{figure*}
  \centering
  \includegraphics[width=0.715\linewidth]{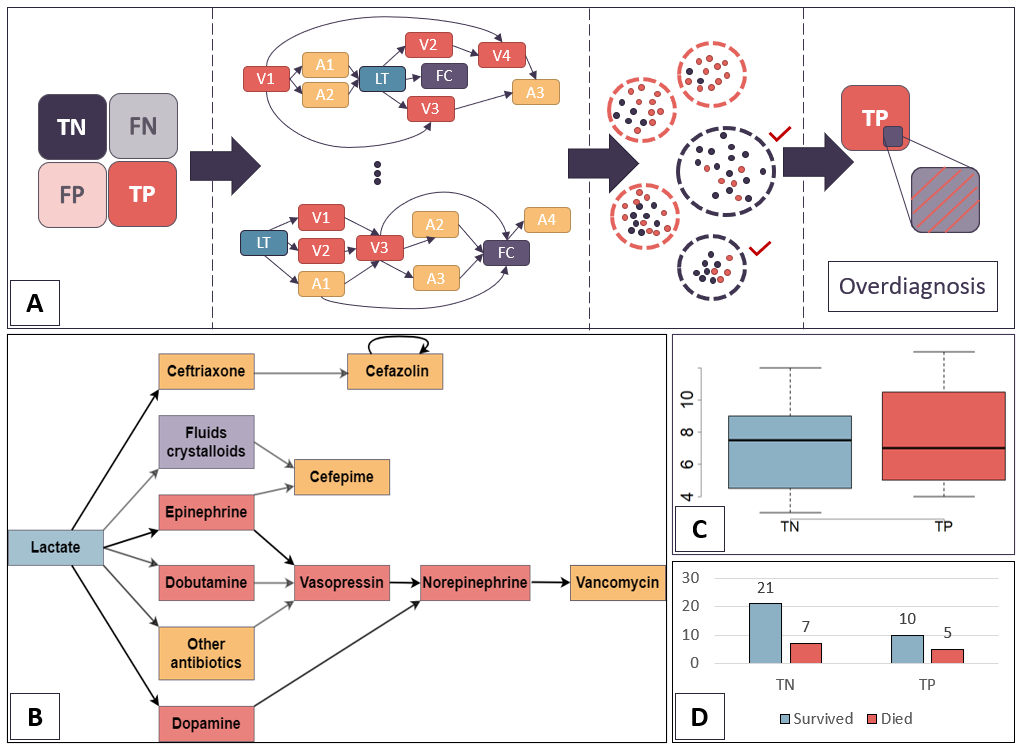}
  \caption{\textbf{A.}	Workflow of the study: building sepsis prediction model, creating, and clustering clinical trajectories, characterizing potential cases of overdiagnosis among true positive patients.
\textbf{B.}	Medical trajectory for cluster 6 obtained from ProM 6.1.
\textbf{C.}	Comparison of SOFA scores between sepsis positive and negative groups in cluster 6. Wilcoxon test shows that difference is statistically insignificant (P > 0.05).
\textbf{D.}	Death rate in cluster 6. Test of Equal or Given Proportions indicated p-value > 0.05.}
  \label{fig:teaser}
\end{figure*}

\section{Methods}

\subsection{Data collection and inclusion criteria}
We chose sepsis as a test case for this study since the likelihood of overdiagnosis is high~\cite{ref5, ref6, ref7}, sepsis is managed in critical care settings where large amounts of data are collected, and cases are fully resolved during hospitalization, thus allowing to quickly determine disease trajectories and prognosis.

For the purpose of this study, we used the MIMIC-IV ~\cite{ref11} database. MIMIC-IV contains de-identified clinical data from patients admitted to the Beth Israel Deaconess Medical Center (BIDMC, Boston Massachusetts, USA) between 2008 – 2019. This dataset includes clinical and administrative information such as patient demographics, transfers, vital signs, laboratory tests, procedures, medications, and diagnoses. 

We included adult patients without an early suspicion of infection, defined as the absence of ordered cultures or antibiotic prescriptions during the first 24 hours of being admitted to the ICU. To avoid data duplication, we used information collected during the first ICU stay if there were more than one ICU stays during one hospital admission. The final cohort contained 39,216 ICU stays with 1,988 positive and 37,228  negative cases.

\subsection{Method description}

The overall workflow of the study consists of the following steps and is illustrated in (Figure~\ref{fig:teaser}A):

\begin{itemize}
\item {Sepsis prediction model}: we built a sepsis prediction model employing supervised machine learning using \textit{sepsis-3} as the ground truth as described in~\cite{ref18}. For further analysis we kept only patients with true positive and true negative labels.
\item {Clinical Trajectories}: we created clinical trajectories for each patient using relevant activities from the surviving sepsis campaign guideline~\cite{ref8}. 
\item {Trajectory Clusters}: we then clustered clinical trajectories using active trace clustering (ActiTraC) algorithm~\cite{ref14}.
\item {Case Analysis}: to identify potential overdiagnosis, we assessed true positive patients with disease trajectories that clustered with true negative ones.
\end{itemize}




\subsection{Sepsis prediction model}

Data collected during the first 24 hours of a patient’s stay in ICU were used as an evidence for training of the sepsis prediction model. Four groups of features were used for the modelling: vital signs, laboratory test results, SOFA-related features and scores, and medication administration. Based on greedy feature selection, 13 best features were included into the final model.

Since the final cohort has significant class imbalance (there are only 4.6\% of sepsis positive cases), an undersampling strategy was used for building a balanced dataset. Experiments showed that the best performance was achieved by employing Light Gradient Boosting Machine (LightGBM) algorithm.

\subsection{Event Log and Clinical Trajectories}

We created an event log for patients classified as true positives or true negatives by the predictive model and the ground truth and whose trajectories contained at least three different events. We chose three events as a minimum length to provide enough information for analysis. The dataset for clustering consisted of 1,421 different traces (680 sepsis positive and 741 sepsis negative) with a mean length of 4 events per trace. Total number of events was 13: 'lactate' test, 'fluids crystalloids', 5 events associated with vasopressor therapy: 'norepinephrine', 'epinephrine', 'vasopressin', 'dopamine', 'dobutamine' and 6 events related to antibiotic administration: 'vancomycin', 'other antibiotics', 'cefepime', 'piperacillin-tazobactam', 'ceftriaxone', 'cefazolin'.

\subsection{Clustering}

To cluster trajectories, we used active trace clustering (ActiTraC) plugin in ProM 6.1~\cite{ref15}. ActiTraC  finds an optimal distribution of execution traces over a given number of clusters, maximizing combined accuracy of the associated process models, also employing principles of active learning. This algorithm has two selective sampling strategies~\cite{ref14}. For this work we chose ‘Distance based selective sampling’ strategy with MRA-based (maximum repeat alphabet) euclidean distance function as defined in~\cite{ref16}. We employed an iterative approach using overall accuracy of the process model as evaluation metric to define parameters for the ActiTraC algorithm (target ICS-fitness was 95\%, maximum number of clusters was 24, minimum cluster size was 4 traces and window size for selective sampling was 0.5).

\section{Results}

Our balanced sepsis prediction model achieved an AUROC of 0.86 and MCC of 0.57, under cross validation, which was consistent with performance on a blind test (AUROC=0.86 and MCC=0.57), demonstrating generalisation capabilities. For an imbalanced blind dataset, the model showed an AUROC of 0.86 and MCC of 0.28. 

After constructing the event log for patients correctly labelled by the sepsis prediction model and applying the clustering algorithm, we got the following results. For the event log of 1,421 cases, 24 clusters were obtained with the average cluster size of 59 traces. The largest and the smallest clusters consisted of 374 (26.31\%) and 6 (0,42\%) traces, respectively. All remaining 288 (20.26\%) unassigned traces were put into a separated cluster. Most clusters had size of 16-72 traces (1-5\% of the event log).

We then evaluated clustering composition. Positive cases are distributed as follows: 10 clusters (41\%) among 24 have less than 50\% of sepsis positive instances. There are two clusters with more than 75\% of positive incidents and one cluster that consists of sepsis negative instances only. Clusters with relatively small shares of sepsis cases presented vasopressor medication, blood test on lactate level and antibiotics administration in primary sequencies of their trajectories. For clusters with high proportion of sepsis cases, antibiotics administration is the first or second action in a trajectory. We expect that potential cases of overdiagnosis to be in clusters skewed towards sepsis negative subgroup and that contain at least 10 sepsis positive cases.

To evaluate similarity of patients in a cluster we analysed combination of several factors: mortality rate, SOFA scores during the first 24 hours in ICU and discharge location. 

Using Wilcoxon test for a SOFA score and test of Equal or Given Proportions for mortality rate we assessed similarity of patients in terms of outcome in each cluster. Among all clusters with lower than 50\% of sepsis positive cases, two clusters (6 and 12) showed no statistically significant difference in SOFA and death rate with 95\% of confidence (Figure~\ref{fig:teaser} B,C,D). In cluster 6, the proportions of patients in positive and negative groups who were discharged to rehabilitation centres or skilled nursing facilities are roughly the same (18\% and 20\% in each of the locations). Since medical trajectories inside these clusters demonstrated no statistical difference in treatment outcomes, we can expect that cases of overdiagnosis are represented in these two clusters. Using the described method, we found 29 cases of potential overdiagnosis, which are 4.3\% of all cases of sepsis included in the study.

\section{Conclusion}
In this paper we proposed for the first time, a computational approach for preemptive identification of potential cases of overdiagnosis, which showed promising results. The developed method is a strategy to quantify the problem and to improve tool development which, with the dissemination of machine learning prediction models, will become more prominent. This method has the advantage over other approaches as it can be used to inform model engineering and potentially actively mitigate machine learning-induced overdiagnosis. A limitation of our approach is the available metrics to measure treatment outcomes and can be complemented as more data becomes available (\textit{e.g.}, discharge summary). In future works we intend to supplement medical trajectories with larger number of events and take into consideration alternative parameters describing a patient’s recovery and treatment outcomes.


\bibliographystyle{ACM-Reference-Format}
\bibliography{sample-base}


\end{document}